\documentclass{article}
\usepackage[utf8]{inputenc}
\usepackage[margin=1in]{geometry}
\usepackage{amsmath,amssymb}
\usepackage{graphicx, booktabs}
\usepackage{subfigure,bbm}
\usepackage{multirow}
\usepackage{color, colortbl}
\definecolor{Gray}{gray}{0.85}
\usepackage[table,xcdraw]{xcolor}
\usepackage{floatrow, mathtools}
\usepackage{nameref}
\newfloatcommand{capbtabbox}{table}[][\FBwidth]

\newcommand{\y}{\mathbf{y}}
\newcommand{\x}{\mathbf{x}}
\newcommand{\res}{\mathbf{r}}
\newcommand{\yh}{\hat{\mathbf{y}}}
\newcommand{\ls}{\mathcal{L}}
\newcommand{\ta}{\mathrm{\theta}}
\newcommand{\de}{\mathrm{\delta}}
\newcommand{\ph}{\mathrm{\phi}}

\title{\bf Designing Accurate Emulators for Scientific Processes using Calibration-Driven Deep Models}

\author{Jayaraman J. Thiagarajan$^{\dagger *}$, Bindya Venkatesh$^+$, Rushil Anirudh$^{\dagger}$, \\
	Peer-Timo Bremer$^{\dagger}$, Jim Gaffney$^{\dagger}$, Gemma Anderson$^{\dagger}$, Brian Spears$^{\dagger}$ \\ \\
	$^{\dagger}$Lawrence Livermore National Laboratory, $^{+}$Arizona State University
}

\author{Jayaraman J. Thiagarajan$^*$\\
	Lawrence Livermore National Laboratory \\
	Email: jjayaram@llnl.gov \\
	\\
	 Bindya Venkatesh \\
	Arizona State University \\
	Email: bindya.venkatesh@asu.edu
	\\ \\
	Rushil Anirudh\\
	Lawrence Livermore National Laboratory \\
	Email: anirudh1@llnl.gov \\
	\\
	Peer-Timo Bremer \\
	Lawrence Livermore National Laboratory \\
	Email: bremer5@llnl.gov \\
	\\
	Jim Gaffney \\
	Lawrence Livermore National Laboratory \\
	Email: gaffney3@llnl.gov \\
	\\
	Gemma Anderson \\
		Lawrence Livermore National Laboratory \\
	Email: anderson276@llnl.gov \\
	\\
	Brian Spears \\
	Lawrence Livermore National Laboratory \\
	Email: spears9@llnl.gov
}

\date{}

\begin{document}
	\maketitle
	
		\newpage
	\begin{abstract}
			\noindent Predictive models that accurately emulate complex scientific processes can achieve exponential speed-ups over numerical simulators or experiments, and at the same time provide surrogates for improving the subsequent analysis. Consequently, there is a recent surge in utilizing modern machine learning (ML) methods, such as deep neural networks, to build data-driven emulators. While the majority of existing efforts has focused on tailoring off-the-shelf ML solutions to better suit the scientific problem at hand, we study an often overlooked, yet important, problem of choosing loss functions to measure the discrepancy between observed data and the predictions from a model. Due to lack of better priors on the expected residual structure, in practice, simple choices such as the mean squared error and the mean absolute error are made. However, the inherent symmetric noise assumption made by these loss functions makes them inappropriate in cases where the data is heterogeneous or when the noise distribution is asymmetric. We propose Learn-by-Calibrating (LbC), a novel deep learning approach based on interval calibration for designing emulators in scientific applications, that are effective even with heterogeneous data and are robust to outliers. Using a large suite of use-cases, we show that LbC provides significant improvements in generalization error over widely-adopted loss function choices, achieves high-quality emulators even in small data regimes and more importantly, recovers the inherent noise structure without any explicit priors.
	\end{abstract}

	\section*{Introduction}	
    \noindent Building functional relationships between a collection of observed \textit{input} variables $\x = \{x_1, \cdots, x_d\}$ and a \textit{response} variable $\y$ is a central problem in scientific applications -- examples range from estimating the future state of a molecular dynamics simulation~\cite{butler2018machine} to searching for exotic particles in high-energy physics~\cite{baldi2014searching} and detecting the likelihood of disease progression in a patient~\cite{esteva2019guide}.
Emulating complex scientific processes using computationally efficient predictive models can achieve exponential speed-ups over traditional numerical simulators or conducting actual experiments, and more importantly provides surrogates for improving the subsequent analysis steps such as inverse modeling, experiment design, etc.
Commonly referred to as supervised learning in the machine learning literature, the goal here is to infer the function $f: \x \mapsto \y$ using a training sample $\{\x_i, \y_i\}_{i=1}^n$, such that the expected discrepancy between $\y$ and $f(\x)$, typically measured using a loss function $\mathcal{L}(\y, f(\x))$, is minimized over the joint distribution $p(\x, \y)$. 

With the availability of modern representation learning methods that can handle complex, multi-variate datatypes, the response variable $\y$ can now correspond to quantities ranging from a collection of scalars, to images, multi-variate time-series measurements, and even symbolic expressions, or combinations thereof.
In particular, the success of deep neural networks in approximating scientific processes involving different types of response variables has generated significant research interest towards improving the accuracy and reliability of emulators~\cite{anirudh2019improved,zhu2018bayesian,paganini2018calogan,peurifoy2018nanophotonic}.
This includes the large body of recent works on incorporating known scientific priors as constraints into predictive modeling~\cite{zhu2019physics}, designing custom neural network architectures that can systematically preserve the underlying symmetries~\cite{cranmer2020lagrangian}, integrating uncertainty quantification methodologies to improve model reliability~\cite{zhu2018bayesian}, and devising novel learning techniques that can handle the inherent data challenges in scientific problems (e.g. small data, under-determined systems)~\cite{anirudh2019improved}. However, a fundamental, yet often overlooked, aspect of this problem is the choice of the loss function $\mathcal{L}$. Denoting $\y = f(\x) + \mathrm{n}$, where $ \mathrm{n}$ denotes the inherent noise in the observed data, the loss function used to measure the discrepancy $\y - f(\x)$ is directly linked to the assumptions made on the noise distribution.

Despite the importance of $\mathcal{L}$ in determining the fidelity of $f$, in practice, simple metrics, such as the $\ell_2$-metric, $||\y - f(\x)||_2$, are used, mostly for convenience but also due to lack of priors on the distribution of residuals. Especially for complex, over-parameterized models such as deep neural networks, it is unclear how well $f$ is able to fit the training data and how errors might be structured, and hence in practice an MSE style penalty on the error is often considered to be an appropriate ``null-hypothesis''. However, this disregards the inherent characteristics of the training data and more importantly the fact that choosing a metric implicitly defines a prior for $\mathrm{n}$. Yet appropriately accounting for noise is crucial to robustly estimate $f$ and to create high-fidelity predictions for unseen data. In the case of the $\ell_2$-metric one implicitly assumes the residuals  $\res = (\y - f(\x))$ to follow a Gaussian distribution. This leads to the optimal estimate in the maximum likelihood sense to be the conditional mean $E(\y|\x)$, which justifies the use of the mean squared error (MSE) in practice.
However, this assumption can be easily violated in real-world data.
For example, the $\ell_2$ metric is known to be susceptible to outliers~\cite{huber2004robust} and cannot handle fast state dynamics such as jumps in the state values~\cite{ohlsson2012smoothed}.
A potential solution is to resort to other symmetric loss functions, e.g. Huber~\cite{huber2004robust} or the Vapnik's $\epsilon-$sensitive loss~\cite{vapnik1998statistical}, that are known to be more robust. However, even those variants can be insufficient when data is more heterogenous, for example, due to heteroscedastic variance or other forms of non-location-scale covariate effects~\cite{wang2012quantile}. With heterogeneous data, merely estimating the conditional mean is insufficient, as estimates of the standard errors are often biased.
This has led to the design of different parameterized, asymmetric loss functions, e.g. quantile~\cite{wang2012quantile} or quantile Huber~\cite{aravkin2013sparse}, that enable one to explore the entire conditional distribution of the response variable  $p(\y|\x)$ instead of only the conditional mean.
Furthermore, allowing the loss function be asymmetric provides one the flexibility to penalize positive and negative components of the residual differently~\cite{aravkin2013sparse}.
For example, quantile regression uses the following loss function:
\begin{equation}
\ls_{q}^{\tau}(\y, f(\x)) \coloneqq \sum_{i=1}^n \max\bigg[\tau(\hat{y}_i - y_i), (\tau-1)(\hat{y}_i - y_i)\bigg]
\label{eqn:quantile}
\end{equation}where $\tau \in [0,1]$ is the quantile parameter. In this expression, the first term will be positive and dominate when over predicting, $\hat{y}_i > y_i$, and the second term will dominate when under-predicting, $\hat{y}_i < y_i$. The larger the value of $\tau$, the more over-predictions are penalized compared to under-predictions, thus enabling asymmetric noise modeling. When $\tau=0.5$, under-prediction and over-prediction will be penalized by the same factor, and hence this expression simplifies to the standard $\ell_1$ loss. Though quantile regression has been found to be effective in handling heterogeneous data and being robust to outliers, determining the appropriate quantile parameter $\tau$ that reflects the expected degree of asymmetry in the distribution of residuals is challenging.
This becomes even more intractable when the response variable $\y$ is multi-variate, and one needs to determine the parameter $\tau$ for each of the response variables. 

\begin{figure}[t]
	\centering
	\includegraphics[width = 0.99\linewidth]{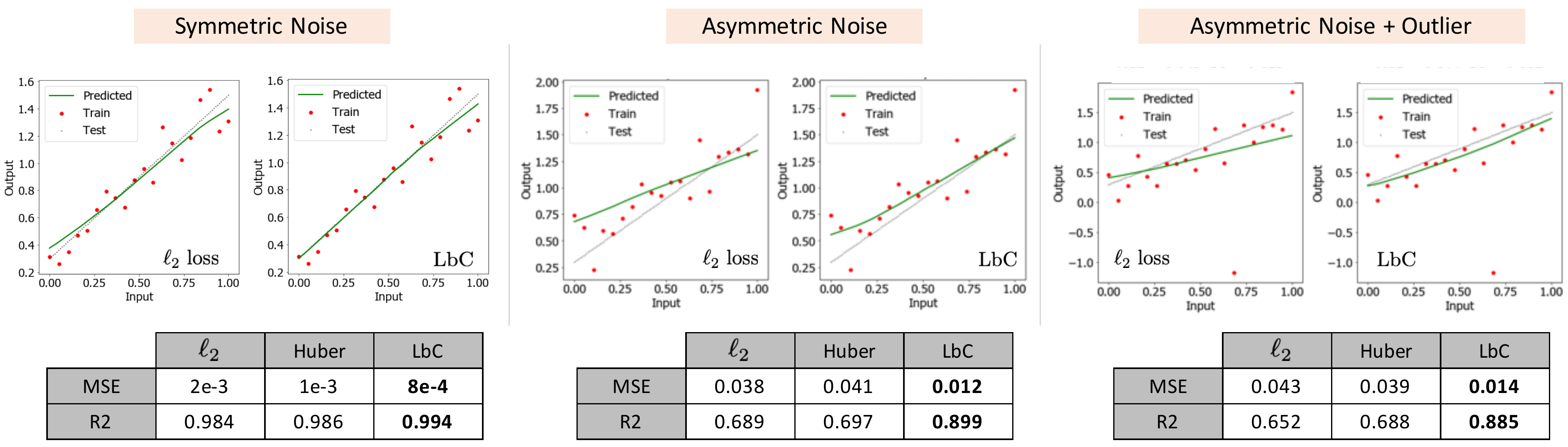}
	\caption{Comparing the behavior of models trained using symmetric losses (e.g. $\ell_2$, Huber) and the proposed Learn-by-Calibrating (LbC) strategy using a simple $1-$D regression experiment. When the noise model for the observed data is symmetric (Gaussian in this case), even the standard MSE loss can recover the true function. However, when the noise model is asymmetric (positive skew), symmetric losses lead to poor approximations. In contrast, LbC can produce higher-fidelity predictions by not enforcing a symmetric residual structure. Finally, when there are outliers in addition to an asymmetric (negative skew) noise model, the non-robustness of the squared error metric becomes clearly evident, while LbC is found to be robust.}
		\label{fig:demo}
\end{figure}

\paragraph{Proposed Work.}
In this paper, we present \textit{Learn-by-Calibrating} (LbC), a non-parametric approach based on interval calibration for building emulators in scientific applications, that are effective even with heterogeneous data and are robust to outliers. The notion of interval calibration comes from the uncertainty quantification literature~\cite{thiagarajan2019building,thiagarajan2019learnbycalibrating} and can be formally defined as follows: Let us assume that the model $f$ is designed to produce prediction intervals, in lieu of simple point estimates, for the response $\y$, i.e., $[\hat{\y} - \mathrm{\delta}^l, \hat{\y} + \mathrm{\delta}^u]$. While the point estimate is a random variable, an interval estimate is a random interval which has a certain probability of containing a value. Suppose that the likelihood for the true response $\y$ to be contained in the prediction interval is $p(\hat{\y} - \mathrm{\delta}^l \leq \y  \leq \hat{\y} + \mathrm{\delta}^u)$, the intervals are considered to be well calibrated if the likelihood matches the expected confidence level. For a confidence level $\alpha$, we expect the interval to contain the true response for $100\times \alpha\%$ of realizations from $p(\x)$. Though calibration has been conventionally used for evaluating and correcting uncertainty estimators, this paper advocates for utilizing calibration as a training objective in regression models. More specifically, LbC uses two separate modules, implemented as neural networks, to produce point estimates and intervals respectively for the response variable, and poses a bi-level optimization problem to solve for the parameters of both the networks. This eliminates the need to construct priors on the expected residual structure and makes it applicable to both homogeneous and heterogeneous data. Furthermore, by effectively recovering the inherent noise structure, LbC leads to highly robust models.  Figure~\ref{fig:demo} provides an illustration with a simple $1-$D regression experiment, where we find that LbC is consistently superior to the widely adopted $\ell_2$ and Huber loss functions, under both symmetric and asymmetric noise models, as well as in the presence of outliers.

Note that the evaluation metric in each of the examples (and throughout the paper) remains the traditional MSE and the R-squared (R2) statistic. The only difference is the loss function used during training. In the examples of Figure~\ref{fig:demo} with synthetic data, we use a single layer neural network with $100$ neurons and ReLU (rectified linear units) non-linear activation to fit the training data. Interestingly, even though the baseline explicitly minimizes the MSE objective during training, while LbC does not, LbC leads to significantly lower MSE error on the validation data (or higher R2). We attribute this apparent discrepancy to the data-driven noise model of the LbC objective which generalizes better to unseen data. 
%

\paragraph{Results.}We evaluated the proposed approach using a large suite of use-cases, which require the design of accurate emulators for the underlying scientific processes, namely: (i) predicting the critical temperature of a superconductor based on its chemical formula~\cite{hamidieh2018data}; (ii) airfoil self-noise estimation in aeronautical systems~\cite{lopez2008neural}; (iii) estimating compressive strength of concrete based on its material composition~\cite{yeh2006analysis}; (iv) approximating a decentralized smart grid control simulation that characterizes the stability of an energy grid~\cite{arzamasov2018towards}; (v) mimicking the clinical scoring process from biomedical measurements in Parkinsons patients~\cite{tsanas2009accurate}; (vi) emulating a semi-analytical 1D simulator (JAG) for inertial confinement fusion that produces multiple diagnostic scalars~\cite{JAG_LLNL}; (vii) emulating a 2D simulator for inertial confinement fusion that produces multi-modal outputs; and (viii) emulating a reservoir simulator that provides estimates for oil and water production over time~\cite{lun2012procedure}. These benchmarks represent a broad range of real-world scenarios including different sample sizes, varying input dimensionality and the need to handle response variable types ranging from single/multiple scalar quantities and  multi-variate time-series measurements to multi-modal outputs. 

Our empirical studies clearly demonstrate the effectiveness of calibration-based training in inferring high-fidelity functional approximations to complex scientific processes. We find that it consistently outperforms several state-of-the-art baselines including different variants of deep neural networks and ensemble techniques, such as random forests and gradient boosting machines, trained with the widely adopted MSE and Huber loss functions. Furthermore, when compared to deep networks trained with the symmetric losses, we find that LbC can operate reliably even in small data regimes (as low as $1000$), producing higher quality models than even ensemble methods. Another interesting observation is that, on all benchmark problems considered, the distribution of residuals obtained using LbC are skewed and heavy-tailed, i.e., non-Gaussian, and this explains its advantage over standard symmetric losses. Surrogates designed using LbC can effectively emulate even sophisticated simulators such as the ICF Hyrda with a multi-modal response, and the reservoir simulator with a multi-variate time-series response. In summary, LbC is a simple, yet powerful, approach to design emulators that are robust, reflect the inherent data characteristics, generalize well to unseen samples and reliably replace accurate (expensive) simulators in scientific workflows.

\begin{table}
	\centering
	\caption{Description of the use-cases considered in our study for benchmarking the proposed approach.}
	\setlength{\extrarowheight}{0pt}
	\addtolength{\extrarowheight}{\aboverulesep}
	\addtolength{\extrarowheight}{\belowrulesep}
	\setlength{\aboverulesep}{0pt}
	\setlength{\belowrulesep}{0pt}
	\begin{tabular}{|l|ccc|}
		\toprule
		\rowcolor[rgb]{0.753,0.753,0.753}\multicolumn{1}{|c}{\textbf{Test Case}}&\# \textbf{Inputs}&\# \textbf{Outputs}&\multicolumn{1}{c|}{\# \textbf{Samples}}\\
		\hline \hline
		Superconductivity&81&1&21,263\\
		Airfoil Self-Noise&5&1&1,503\\
		Concrete&8&1&1,030\\
		Electric Grid Stability&12&1&10,000\\
		Parkinsons&16&1&5,875\\
		ICF JAG (Scalars)&5&15&10,000\\
		ICF Hydra (Scalars)&9&28&92,965\\
		ICF Hydra (Multi)&9&32&92,965\\
		Reservoir Model&14&14&2,000\\
		\bottomrule
	\end{tabular}
	\label{table:data}
\end{table}


	\section*{Data Description}
	We consider a large suite of scientific problems and design emulators using state-of-the-art predictive modeling techniques. The primary focus of this study is to investigate the impact of using a calibration-driven training objective, in lieu of widely-adopted loss functions, on the quality of emulators. The problems that we consider encompass a broad range of applications, response types and data sizes, and  enable us to rigorously benchmark the proposed approach. Table~\ref{table:data} provides a description of datasets used in each of the use-cases.

\paragraph{Superconductivity.} Superconducting materials, which conduct current with zero resistance, are an integral part of Magnetic Resonance Imaging(MRI) systems and utilized for designing coils to maintain high magnetic fields in particle accelerators. A superconductor exhibits its inherent zero-resistance property only at or below its critical temperature ($T_c$). Developing scientific theory or a model to predict $T_c$ has been an open problem, since its discovery in 1911, and hence empirical rules are used in practice. For example, it has been assumed that the number of available valence electrons per atom is related to $T_c$, though there is recent evidence that this rule can be violated~\cite{conder2016second}. Hence, building statistical predictive models, based on a superconductor's chemical formula, has become an effective alternative~\cite{hamidieh2018data}. This dataset relates $81$ elemental properties of each superconductor to the critical temperature on a total of  $21,263$ samples.

\begin{table*}[t]
	\centering
	\caption{Average Mean Squared Error(MSE) obtained over 5-fold cross validation on each of the use-cases using emulators designed with different machine learning approaches.The best performance in each case is denoted in bold.}
	\setlength{\extrarowheight}{0pt}
	\addtolength{\extrarowheight}{\aboverulesep}
	\addtolength{\extrarowheight}{\belowrulesep}
	\setlength{\aboverulesep}{0pt}
	\setlength{\belowrulesep}{0pt}
	\begin{tabular}{|l|cccccc|}
		\toprule
		\rowcolor[HTML]{C0C0C0}
		\multicolumn{1}{|c|}{\cellcolor[HTML]{C0C0C0}}&\multicolumn{6}{c|}{\cellcolor[HTML]{C0C0C0}Methods}\\\cline{2-7}
		\rowcolor[HTML]{EFEFEF}
		\multicolumn{1}{|c|}{\multirow{-2}{*}{\cellcolor[HTML]{C0C0C0}Test Case}}&\multicolumn{1}{c|}{\cellcolor[HTML]{EFEFEF}RF}&\multicolumn{1}{c|}{\cellcolor[HTML]{EFEFEF}GBT (L2)}&\multicolumn{1}{c|}{\cellcolor[HTML]{EFEFEF}GBT (h)}&\multicolumn{1}{c|}{\cellcolor[HTML]{EFEFEF}DNN}&\multicolumn{1}{c|}{\cellcolor[HTML]{EFEFEF}DNN (drp)}&\multicolumn{1}{c|}{\cellcolor[HTML]{EFEFEF}Proposed}\\\hline
		Electric Grid Stability&0.004&0.0056&0.0055&0.0021&0.00128&\textbf{0.00103}\\
		Concrete&0.02&0.015&0.0152&0.0127&0.0125&\textbf{0.011}\\
		Parkinsons&0.0655&0.062&0.0629&0.0749&0.0638&\textbf{0.047}\\
		Superconductivity&\textbf{0.0053}&0.0066&0.0066&0.0087&0.0079&0.0057\\
		Airfoil Self-Noise&0.0095&0.0125&0.0128&0.0155&0.0121&\textbf{0.0085}\\
		ICF JAG (Scalars)&1.2E-04&3.2E-04&4.3E-04&1.3E-04&6.5E-05&\textbf{4.9E-05}\\
		ICF Hydra (Scalars)&6.2E-04&1.5E-03&1.7E-03&2.5E-04&1.9E-04&\textbf{1.4E-04}\\
		ICF Hydra (Multi)&0.0021&0.006&0.0061&0.001&6.2E-04&\textbf{3.1E-04}\\
		Reservoir Model&0.0032&0.0032&0.0036&0.0025&0.0016&\textbf{0.0014}\\
		\bottomrule
	\end{tabular}
\label{table:mse}
\end{table*}

\begin{table*}[!t]
	\centering
	\caption{Average R-squared statistic (R2) obtained over 5-fold cross validation on each of the use-cases using emulators designed with different machine learning approaches.The best performance in each case is denoted in bold.}
	\setlength{\extrarowheight}{0pt}
	\addtolength{\extrarowheight}{\aboverulesep}
	\addtolength{\extrarowheight}{\belowrulesep}
	\setlength{\aboverulesep}{0pt}
	\setlength{\belowrulesep}{0pt}
	\begin{tabular}{|l|cccccc|}
		\toprule
		\rowcolor[HTML]{C0C0C0}
		\multicolumn{1}{|c|}{\cellcolor[HTML]{C0C0C0}}&\multicolumn{6}{c|}{\cellcolor[HTML]{C0C0C0}Methods}\\\cline{2-7}
		\rowcolor[HTML]{EFEFEF}
		\multicolumn{1}{|c|}{\multirow{-2}{*}{\cellcolor[HTML]{C0C0C0}Test Case}}&\multicolumn{1}{c|}{\cellcolor[HTML]{EFEFEF}RF}&\multicolumn{1}{c|}{\cellcolor[HTML]{EFEFEF}GBT (L2)}&\multicolumn{1}{c|}{\cellcolor[HTML]{EFEFEF}GBT (h)}&\multicolumn{1}{c|}{\cellcolor[HTML]{EFEFEF}DNN}&\multicolumn{1}{c|}{\cellcolor[HTML]{EFEFEF}DNN (drp)}&\multicolumn{1}{c|}{\cellcolor[HTML]{EFEFEF}Proposed}\\\hline
		Electric Grid Stability&0.89&0.85&0.851&0.94&0.96&\textbf{0.972}\\
		Concrete&0.31&0.49&0.5&0.63&0.61&\textbf{0.68}\\
		Parkinsons&0.62&0.55&0.58&0.61&0.61&\textbf{0.73}\\
		Superconductivity&\textbf{0.74}&0.65&0.66&0.57&0.62&0.72\\
		Airfoil Self-Noise&0.71&0.62&0.63&0.63&0.64&\textbf{0.74}\\
		ICF JAG (Scalars)&0.995&0.988&0.983&0.975&0.991&\textbf{0.998}\\
		ICF Hydra (Scalars)&0.88&0.81&0.79&0.88&0.89&\textbf{0.94}\\
		ICF Hydra (Multi)&0.83&0.72&0.73&0.92&0.95&\textbf{0.97}\\
		Reservoir&0.87&0.87&0.85&0.89&0.93&\textbf{0.97}\\
		\bottomrule
	\end{tabular}
\label{table:r2}
\end{table*}

\paragraph{Airfoil Self-Noise.} Controlling the noise generated by an aircraft, in particular the self-noise of the airfoil itself, is essential to improving its efficiency. The self-noise corresponds to the noise generated when the airfoil passes through smooth non-turbulent inflow conditions. The so-called \textit{Brooks} model, a semi-empirical approach for self-noise estimation, has been routinely used over $3$ decades, though it is known to under-predict the noise level in practice. In the recent years, data-driven models are being used instead~\cite{lopez2008neural} and it is crucial to improve the fidelity of such an emulator. This dataset consists of $1503$ cases and $5$ features including the frequency, angle of attack and chord length to predict self-noise.

\paragraph{Concrete.} The key objective of this popular UCI benchmark is to estimate the compressive strength of a concrete, which is known to be a highly non-linear function of its age and material composition. Similar to many other problems in engineering, machine learning approaches have been found to be superior to  heuristic models for estimating the target function~\cite{yeh2006analysis}. This falls under the class of small-data problems, by containing only $1,030$ samples in $8$ dimensions representing the material composition, e.g. amount of cement, fly ash etc.

\paragraph{Electric Grid Stability.} The Decentralized Smart Grid Control (DSGC) system is a recently developed approach for modeling changes in electricity consumption in response to electricity-price changes. A key challenge in this context is to predict the stability, i.e., whether the behavior of participants in response to price changes can destabilize the grid. This dataset contains $10,000$ instances representing local stability analysis of the $4$-node star system, where each instance is described using $12$ different features~\cite{arzamasov2018towards}.

\paragraph{Parkinsons.} Parkinsons is the second most common neurodegenerative disorder after Alzheimers. Though medical intervention can control its progression and  alleviate some of the symptoms, there is no available cure. Consequently, early diagnosis has become a critical step towards improving the patient's quality of life~\cite{tsanas2009accurate}. With the advent of non-invasive monitoring systems in healthcare, its use for early diagnosis in Parkinsons patients has gained significant interest. The goal of this use-case is to predict the severity of disease progression, quantified via the Unified Parkinsons Disease Rating Scale (UPDRS), from speech signals (\textit{vowel phonotations}). The dataset is comprised of $5,875$ patients represented using $16$ different speech features.

\paragraph{ICF JAG.} JAG~\cite{JAG_LLNL} is a semi-analytical $1$D simulator for inertial confinement fusion (ICF), which models a high-fidelity mapping from the process inputs, e.g. target and laser settings, to process outputs, such as the ICF implosion neutron yield. The physics of ICF is predicated on interactions between multiple strongly nonlinear physics mechanisms that  have  multivariate dependence on a large number of controllable parameters. Despite the complicated, non-linear nature of this response, machine learning methods such as deep learning have been showed to produce high-quality emulators~\cite{anirudh2019improved}. This dataset contains $10,000$ samples with $5$ input parameters and $15$ scalar quantities in the response.

\paragraph{ICF Hydra.} Hydra is a $2$D physics code used to simulate capsule implosion experiments~\cite{langer2016hydra}. This has the physics required to simulate NIF capsules including hydrodynamics, radiation transport, heat conduction, fusion reactions, equations of state, and opacities. It consists of over a million lines of code and takes hours to run a single simulation. In terms of sample size, this is a fairly large-scale data with about $93$K simulations, where each sample corresponds to $9$ input parameters and a multi-modal response ($2$ channel X-ray images, $28$ scalar quantities, FNADS). In our experiments, we consider two different variants, one with only the multi-variate scalar response and another with the entire multi-modal response. Following the protocol in~\cite{anirudh2019improved}, in the case of multi-modal responses, we first build an encoder-decoder style neural network that transforms the multi-modal response into a joint latent space of $32$ dimensions and repose the surrogate modeling problem as predicting from the input parameters into the low-dimensional latent space. We can recover the actual response using the decoder model on the predicted latent representations.

\begin{figure*}[!t]
	\centering
	\includegraphics[width=0.99\linewidth]{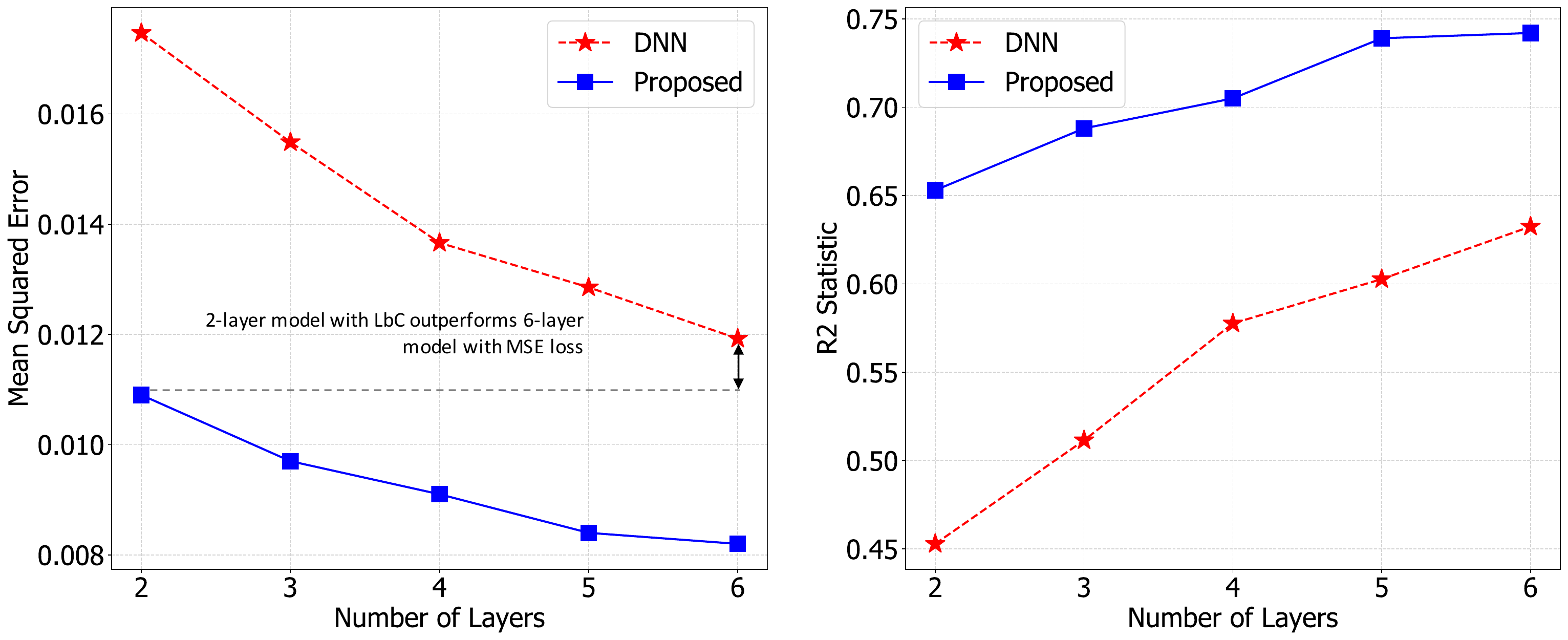}
	\caption{\textit{Airfoil Self-Noise dataset}: Comparing the performance of emulators designed using conventional deep neural networks with MSE as the optimization objective and the proposed approach that utilizes a calibration objective for training deep models. We find that regardless of the complexity of the model (varying depth), the proposed approach produces significantly superior emulators. Though LbC uses an additional network for estimating the intervals during training, at inference time, the predictions are obtained using only the network $f$ whose number of parameters are exactly same as that of the DNN baseline.}
	\label{fig:mse-lbc}
\end{figure*}

\paragraph{Reservoir Model.} This simulator models a two-well waterflood in a reservoir containing two stacked channel complexes. The model represents a deep-water slope channel system, in which sediment is deposited in channel complexes as a river empties into a deep basin. A high-quality surrogate is required to solve the crucial task of \textit{history matching}, an ill-posed inverse problem for calibrating model parameters to real-world measurements. The dataset contains $2000$ simulations with $14$ input parameters and $3$ time histories corresponding to injection pressure, oil and water production rates. Similar to the ICF Hydra case, we use an auto-encoder model to transform the multi-variate time-series response into a $14$-dimensional latent space. Note, we use the network architecture in~\cite{narayanaswamydesigning} for designing the auto-encoder.


	\section*{Results}
	\begin{figure*}[!t]
	\centering
	\includegraphics[width=0.99\linewidth]{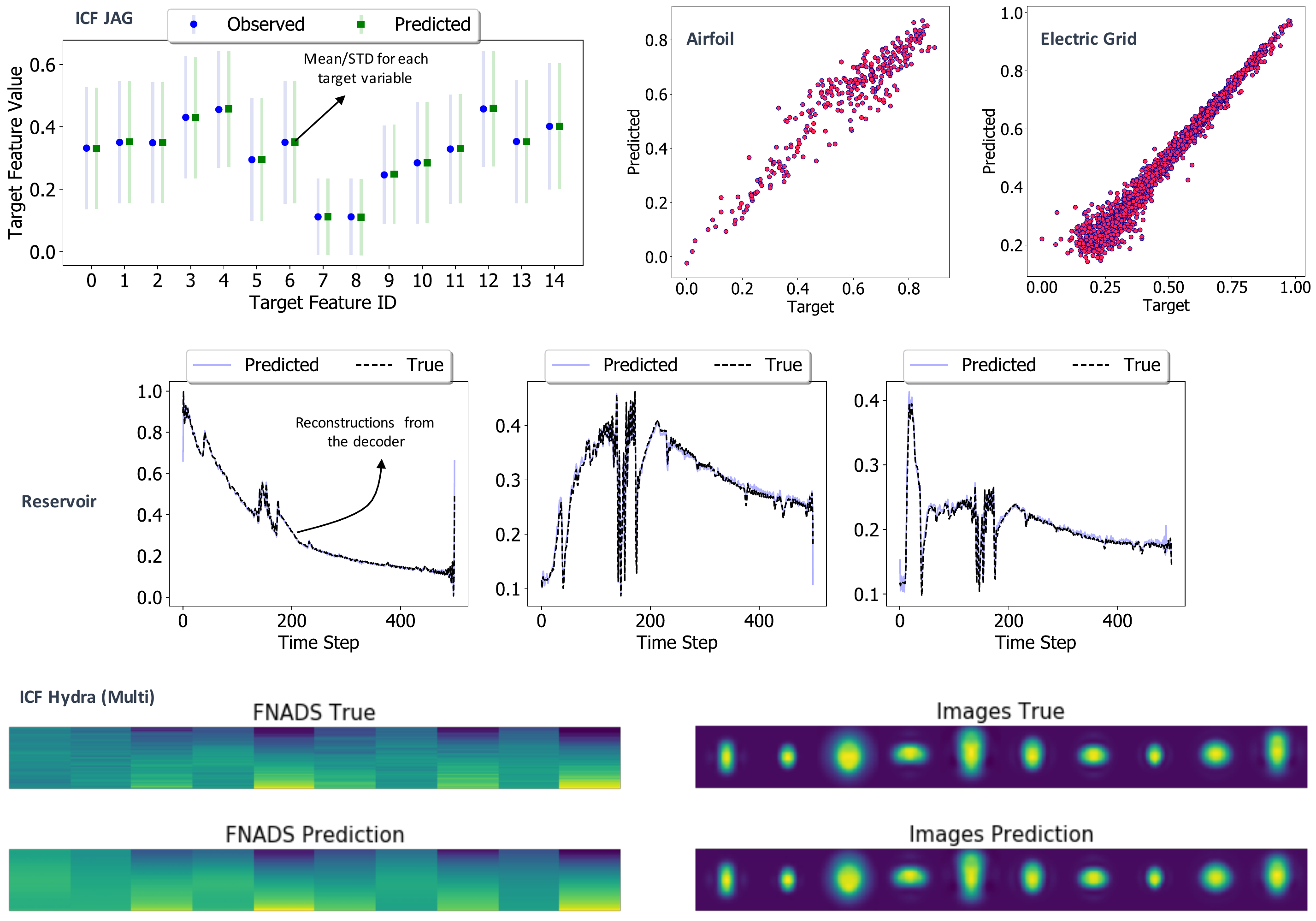}
	\caption{Predictions obtained using the proposed approach on different use-cases. Across scientific datasets of varying dimensionality and complexity, LbC produces high-fidelity emulators that can be reliably used in scientific workflows. }
	\label{fig:preds}
\end{figure*}

\begin{figure*}[t]
	\centering
	\includegraphics[width=0.99\linewidth]{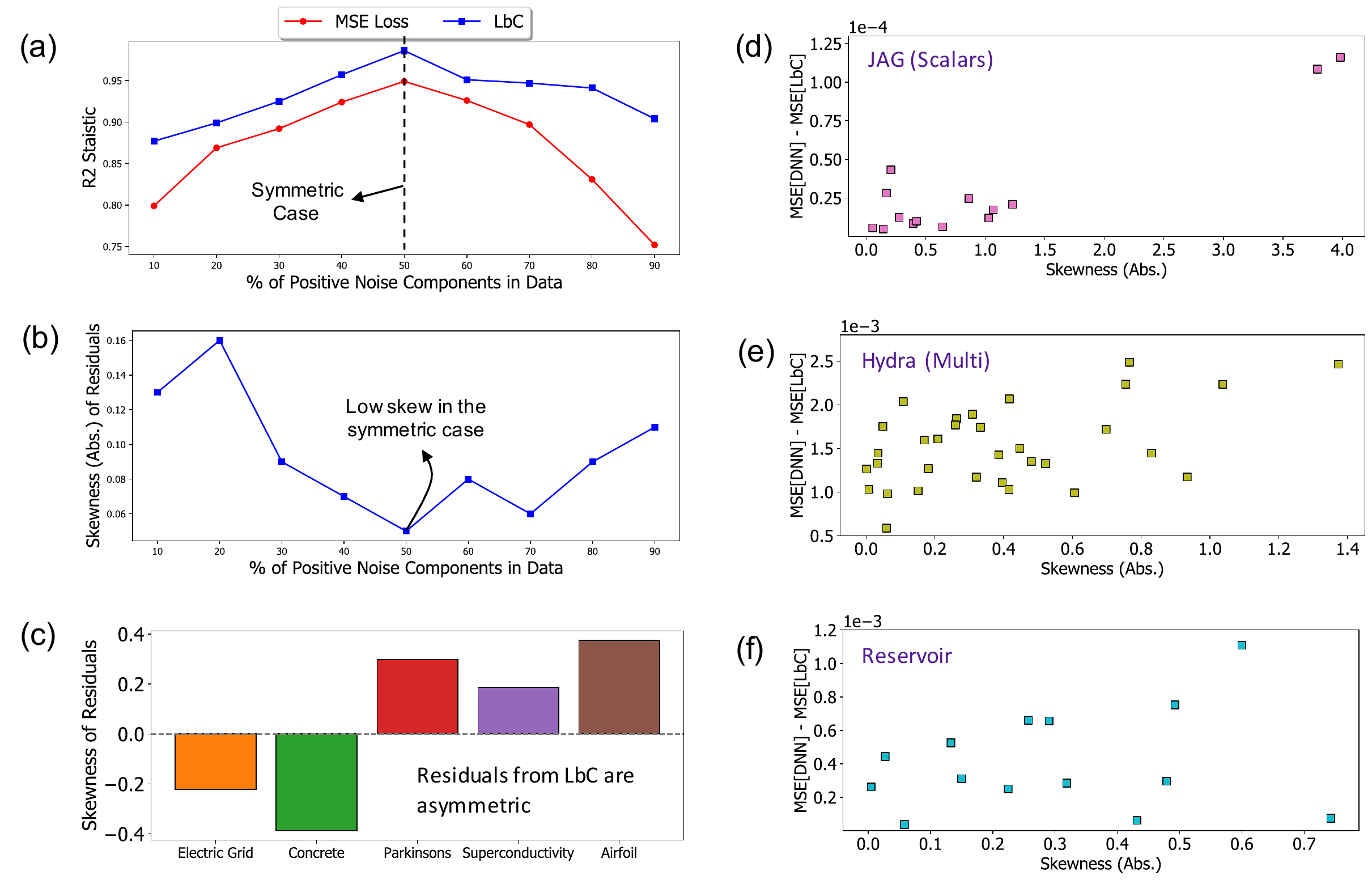}
	\caption{Analysis: Using the synthetic function in Figure \ref{fig:demo}, we find that (a) LbC produces significantly improved generalization at varying levels of asymmetry in the inherent noise structure and (b) the skewness of the residuals from LbC reflect that. For all the test-cased considered in our study, (c) we find that the residuals are highly asymmetric and heavy-tailed. Interestingly, from figures (d)-(f), we observe that, in cases where the performance gains are significant (difference between MSEs of DNN and LbC), the corresponding skewness of the residual distribution is high. This clearly evidences the ability of our approach to reveal the inherent noise structure in the data.}
	\label{fig:residual}
\end{figure*}

We report the performance of LbC, in comparison to several state-of-the-baselines, on the benchmark problems. We use two standard evaluation metrics, namely mean-squared error (lower is better) and the R-squared statistic (R2), which measures the proportion of variance in the response variable that is predictable from the input variable (higher is better). For the empirical analysis, we consider the following baseline methods: 
\begin{itemize}
	\item Random forests (RF) with $100$ decision trees trained using the $\ell_2$ metric;
	\item Gradient boosting machines with $100$ decision trees, trained using the $\ell_2$ (GBT (L2)) and Huber (GBT (h)) loss functions;
	\item Deep neural networks (DNN) with $5$ fully connected layers and a final prediction layer with dimensions corresponding to the response variable (details can be found in the \textit{Methods} section). Note, we used the ReLU non-linear activation after every hidden layer and optimized for minimizing the $\ell_2$ metric; 
	\item A variant of the DNN model, referred as DNN (drp), wherein we introduce dropout-based epistemic uncertainty quantification during training. Dropout is a popular regularization technique that randomly drops hidden units (along with their connections) in a neural network. Following~\cite{gal2016uncertainty}, for each sample, we make $T$ forward passes with the dropout rate set to $\tau$ and obtain the final prediction as the average from the $T$ runs. This is known to produce more robust estimates in regression problems~\cite{thiagarajan2019building}. In our experiments we set $T = 20$ and the dropout rate $\tau = 0.3$.
\end{itemize}To provide statistically meaningful results, we performed $5-$fold cross validation for each of the use-cases and report the average performance. The MSE and R2 scores achieved using the different approaches are reported in Table \ref{table:mse} and Table \ref{table:r2} respectively. We find that LbC consistently produces higher quality emulators in all cases, except for the superconductivity dataset where random forests are marginally better. In terms of the R2 statistic, we find that, LbC achieves an average improvement of $\sim 20\%$ over the popular ensemble methods, regardless of the loss function ($\ell_2$ or Huber) used for training. On the other hand, when compared to the two deep learning baselines, the average improvement in R2 is about $10\%$. Interestingly, with challenging benchmarks such as the superconductivity and airfoil self-noise prediction problems, the neural network based solutions (DNN, DNN (drp)) are inferior to ensemble methods. This can be attributed to the overfitting behavior of over-parameterized neural networks in small data scenarios. In contrast, LbC is highly robust even in those scenarios and produces higher R2 scores (or lower MSE). This is also apparent from the analysis in Figure \ref{fig:mse-lbc}, where we find that even a shallow $2-$layer network with the proposed calibration-driven learning outperforms a standard deep model with $6$ layers. This clearly emphasizes the discrepancy between the true data characteristics and the assumptions placed by the $\ell_2$ loss function. With simulators such as ICF Hydra and the reservoir model, which map to complex response types, our approach makes accurate predictions in the latent space (from the auto-encoder)  and when coupled with the decoder matches the true responses (Figure \ref{fig:preds}) .

In contrast to existing loss functions, LbC does not place any explicit priors on the residual structure and hence it is important to analyze the characteristics of the errors obtained using our approach. Using the synthetic function from Figure~\ref{fig:demo}, we varied the percentage of positive noise components in the observed data ($50\%$ corresponds to the symmetric noise case) and evaluated the prediction performance using the R2-statistic. As showed in Figure~\ref{fig:residual}(a), while LbC outperforms the MSE loss in all cases, with increasing levels of asymmetry the latter approach produces significantly lower quality predictions. This clearly evidences the limitation of using a simple Gaussian assumption or even a more general symmetric noise assumption, when the inherent noise distribution is actually asymmetric. From Figure~\ref{fig:residual}(b), where we plot the skewness of residual distributions, we find that LbC effectively captures the true noise model, thus producing high-fidelity predictors. Furthermore, we make similar observations on the different use-cases (see Figure~\ref{fig:residual}(d)-(f)) -- the maximal performance gains (measured as difference in MSE between the DNN baseline and LbC models with the same network architecture) are obtained when the skewness of the residuals from LbC are large, indicating the insufficiency of MSE loss in modeling real-world scientific data.




	\section*{Methods}
	Learn-by-Calibrating (LbC) is a prior-free approach for training regression models via interval calibration. We begin by assuming that our model produces prediction intervals instead of simple point estimates, i.e., $[\hat{\mathbf{y}} - \mathrm{\delta}^l, \hat{\mathbf{y}} + \mathrm{\delta}^u]$, for an input sample $\mathbf{x}$. More specifically, our model is comprised of two modules $f$ and $g$, implemented as deep neural networks, to produce estimates $\hat{\mathbf{y}} = f(\mathbf{x}; \mathrm{\theta})$ and $(\mathrm{\delta}^l,\mathrm{\delta}^u)  = g(\mathbf{x};\mathrm{\phi})$. We design a bi-level optimization strategy to infer $\mathrm{\theta}$ and $\mathrm{\phi}$, i.e., parameters of the two modules, using observed data $\{(\mathbf{x}_i,\mathbf{y}_i)\}_{i=1}^n$:
\begin{align}
\nonumber &\min_{\mathrm{\theta}} \mathcal{L}_f\bigg(\mathrm{\theta}; \{\mathbf{x}_i, \mathbf{y}_i\}_{i=1}^n, g(\mathrm{\phi}^*)\bigg) \\
&\text{s.t. }  \mathrm{\phi}^* = \arg \min_{\mathrm{\phi}} \mathcal{L}_g\bigg(\mathrm{\phi}; \{\mathbf{x}_i\}_{i=1}^n,f(\mathrm{\theta})\bigg).
\label{eqn:bilevel}
\end{align}Here $\mathcal{L}_f$ and  $\mathcal{L}_g$ are the loss functions for the two modules. In practice, we use an alternating optimization strategy to infer the parameters. LbC utilizes interval calibration from uncertainty quantification to carry out this optimization without placing an explicit prior on the residuals. We attempt to produce prediction intervals that can be calibrated to different confidence levels $\alpha$ and hence the module $g$ needs to estimate $(\mathrm{\delta}^{l,\alpha}, \mathrm{\delta}^{u,\alpha})$ corresponding to each $\alpha$. In our formulation, we use $\alpha \in \mathcal{A}$, $\mathcal{A} = [0.1,0.3,0.5,0.7,0.9,0.99]$. Note that, while the choice of $\mathcal{A}$ is not very sensitive, we find that simultaneously optimizing for confidence levels in the entire range of $[0,1]$ is beneficial. However, considering more fine-grained sampling of $\alpha$'s (e.g. $\{0.05, 0.1, \cdots\}$) did not lead to significant performance gains, but required more training iterations. The loss function $\mathcal{L}_g$ is designed using an empirical calibration metric similar to~\cite{thiagarajan2019building}:
\begin{align}
\nonumber \mathrm{\phi}^* = \arg \min_{\mathrm{\phi}} \mathcal{L}_g = \arg \min_{\mathrm{\phi}} \sum_{\alpha \in \mathcal{A}} & \bigg( \bigg| \alpha - \frac{1}{n} \sum_{i=1}^n \mathbbm{1}[\yh_i - \de_i^{l,\alpha} \leq \y_i \leq \yh_i + \de_i^{u,\alpha} ] \bigg| \\ 
&+ \lambda_1 |\y_i - (\yh_i - \de_i^{l,\alpha})|	+ \lambda_2 |(\yh_i + \de_i^{u,\alpha}) -  \y_i|	\bigg).
\label{eqn:phi}
\end{align}Here, $(\de_i^{l,\alpha},\de_i^{u,\alpha})$ represents the estimated interval for sample index $i$ at confidence level $\alpha$, $\mathbbm{1}$ is an indicator function and $\lambda_1, \lambda_2$ are hyper-parameters (set to $0.05$ in our experiments). The first term measures the discrepancy between the expected confidence level and the likelihood of the true response falling in the estimated interval. Note that the estimates $\yh = f(\x; \ta)$ are obtained using the current state of the parameters $\ta$, and the last two terms are used as regularizers to penalize larger intervals so that trivial solutions are avoided. In practice, we find that such a simultaneous optimization for different $\alpha'$s is challenging and hence we randomly choose a single $\alpha$ from $\mathcal{A}$ in each iteration, based on which the loss $\ls_g$ is computed.

\begin{figure*}[!t]
	\centering
	\includegraphics[width=0.8\linewidth]{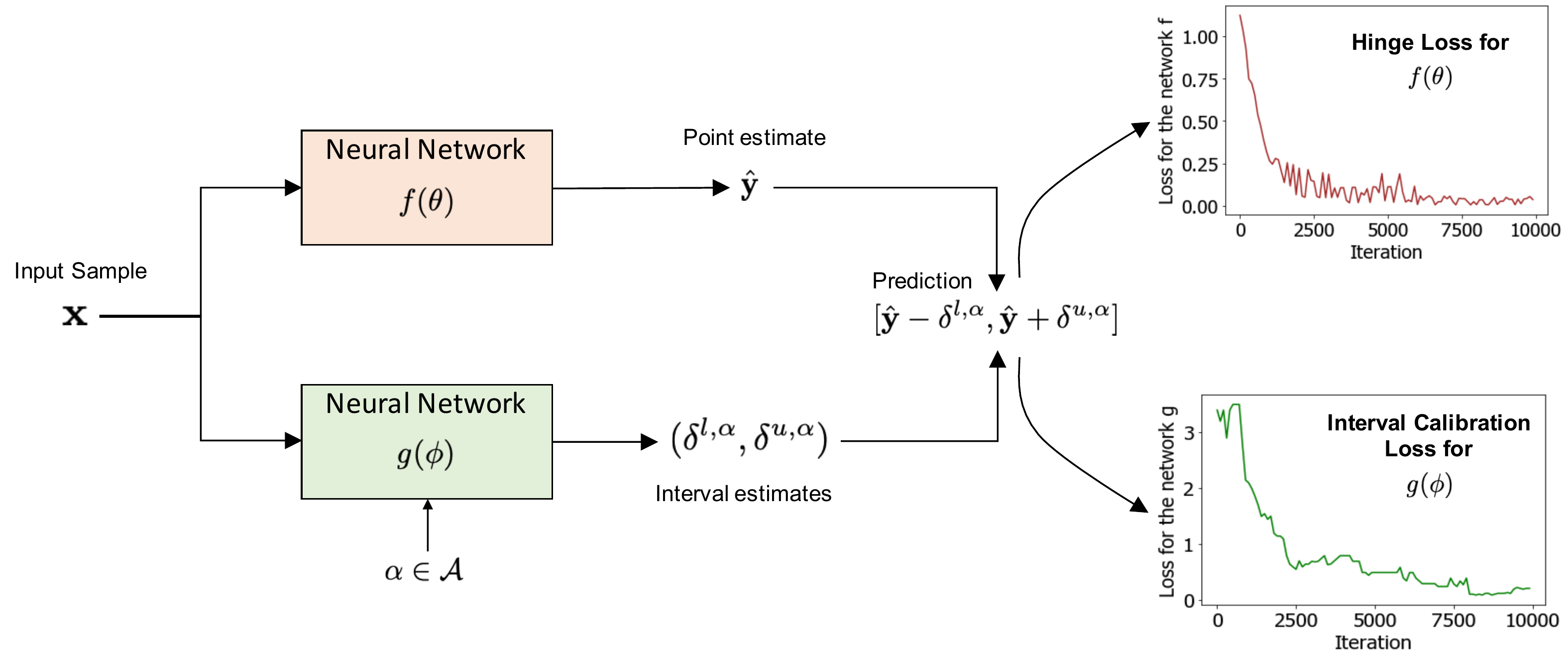}
	\caption{An illustration of the proposed approach, wherein there are two separate networks to obtain point estimates and the intervals respectively. As showed by the convergence plots during training, the two models synergistically optimize for the overall objective of improving the interval calibration.}
	\label{fig:bd}
\end{figure*}

Since LbC relies entirely on calibration, there is no need for explicit discrepancy metrics like $\ell_2$ or Huber for updating the model $f$. Instead, we employ a hinge-loss objective that attempts to adjust the estimate $\yh$ such that the observed likelihood of the true response to be contained in the interval increases:
\begin{equation}
\ta^* = \arg \min_{\ta} \ls_f = \arg \min_{\ta} \sum_{i=1}^n w_i \bigg[ \max(0, (\yh_i - \de_i^{l,\alpha}) -  \y_i + \gamma) + \max(0, \y_i - (\yh_i + \de_i^{u,\alpha})  + \gamma)\bigg].
\label{eqn:theta}
\end{equation}Here, $(\de_i^{l,\alpha},\de_i^{u,\alpha}) = g(\x_i; \ph, \alpha)$ is obtained using the recent state of the parameters $\ph$ and the randomly chosen $\alpha$ in the current iteration, $\gamma$ is a pre-defined threshold (set to $0.05$) and the weights $w_i = (\de_i^{l,\alpha}+\de_i^{u,\alpha})/\sum_j  (\de_j^{l,\alpha}+\de_j^{u,\alpha})$ penalizes samples with larger intervals. Intuitively, the improved estimate $\yh$ can potentially increase the empirical calibration error by achieving a higher likelihood even for smaller $\alpha$. However, in the subsequent step of updating $\ph$, we expect the intervals to become sharper in order to reduce the calibration error. This synergistic optimization process thus leads to superior quality predictions, which we find to be effective regardless of the inherent residual structure. Figure~\ref{fig:bd} illustrates the proposed approach and the convergence curves for the two models $f$ and $g$ obtained for one of the use-cases.

\paragraph{Architecture.} In our implementation, both $f$ and $g$ are implemented as neural networks with fully connected layers and ReLU non-linear activation. For use-cases with at least $5000$ samples, we used $5$ fully connected layers and the number of hidden units fixed at $[64, 128, 512, 256, 32]$ respectively and a final prediction layer. Whereas, we used shallow $3$-layer networks for the smaller datasets ($[64,256,32]$). While the final layer in $f$ corresponds to the dimensionality of the response variable, the final layer in $g$ produces $\delta^l$ and $\delta^u$ estimates for each dimension in $\y$ at every $\alpha \in \mathcal{A}$. The networks were trained using the Adam optimizer with the learning rates for the two modules fixed at $3e-5$ and $1e-4$ respectively and mini-batches of size $64$. The alternating optimization was carried out for about $1000$ iterations.


	\bibliographystyle{IEEEbib}
	\bibliography{refs}

	\section*{Acknowledgments}
	This work was performed under the auspices of the U.S. Department of Energy by Lawrence Livermore National Laboratory under Contract DE-AC52- 07NA27344 and was supported by the LLNL-LDRD
	Program under Project No. 18-SI-002.

	\section*{Disclaimer}	
	This document was prepared as an account of work sponsored by an agency of the United States government. Neither the United States government nor Lawrence Livermore National Security, LLC, nor any of their employees makes any warranty, expressed or implied, or assumes any legal liability or responsibility for the accuracy, completeness, or usefulness of any information, apparatus, product, or process disclosed, or represents that its use would not infringe privately owned rights. Reference herein to any specific commercial product, process, or service by trade name, trademark, manufacturer, or otherwise does not necessarily constitute or imply its endorsement, recommendation, or favoring by the United States government or Lawrence Livermore National Security, LLC. The views and opinions of authors expressed herein do not necessarily state or reflect those of the United States government or Lawrence Livermore National Security, LLC, and shall not be used for advertising or product endorsement purposes.

\end{document}